\documentclass{opt2024} % Include author names

\usepackage{float}
\usepackage{amsmath}

\usepackage{amsfonts}
\usepackage{enumitem}

\title[Applications of fractional calculus in learned optimization]{Applications of fractional calculus in learned optimization}

\optauthor{
\Name{Teodor-Alexandru Szente} \Email{teodor.szente@imar.ro}\\
\addr Institute of Mathematics of the Romanian Academy
\AND
\Name{James Harrison} \Email{jamesharrison@google.com }\\
\addr Google DeepMind
\AND
\Name{Mihai Zanfir} \Email{mihai.zanfir@newton.ro}\\
\addr Newton
\AND
\Name{Cristian Sminchisescu} \Email{sminchisescu@google.com}\\
\addr Google DeepMind, Institute of Mathematics of the Romanian Academy
}

\begin{document}

\maketitle

\begin{abstract}%
Fractional gradient descent has been studied extensively, with a focus on its ability to extend traditional gradient descent methods by incorporating fractional-order derivatives. This approach allows for more flexibility in navigating complex optimization landscapes and offers advantages in certain types of problems, particularly those involving non-linearities and chaotic dynamics.
Yet, the challenge of fine-tuning the fractional order parameters remains unsolved. In this work, we demonstrate that it is possible to train a neural network to predict the order of the gradient effectively.
\end{abstract}

\section{Introduction}
In conventional first order optimization, the target function is typically approximated as locally linear using a Taylor expansion. It is possible to benefit from nonlinear approximations that capture the behavior of the function over a larger vicinity, offering a more accurate representation than local linear approximations. Fractional gradient descent methods were developed to take advantage of such approximations \cite{fractioanalabm, zhiguangfg, shin2021caputo}. As shown in \cite{liangWY2020} they can greatly improve the convergence rate of the gradient descent algorithm in the convex case. These methods rely on the concept of \textit{fractional derivatives}. The fractional derivative can be thought as an "interpolation" between two conventional derivatives. For example, the half derivative (i.e. fractional order $\alpha = 0.5$), denoted as $\frac{d^{0.5}f}{dx^{0.5}}$, represents an interpolation between the function $f$ itself and its first derivative.  However, little insight into determining the optimal fractional order for a specific problem is shown. Adaptive methods have been developed \cite{louvariable2022, Liu2023TheNA} but they depend on additional hyper-parameters (e.g bounds limits, terminal points).
Meanwhile, in the field of learned optimization, improvements have been made for fine tuning expressive optimizers \cite{{li2023variance, harrison2022closer, velo}}. In this paper, we illustrate how learned optimization can be employed to fine-tune the fractional order.
\subsection{Fractional Calculus}
The fractional derivative can be represented as a non-integer extension of the Cauchy formula for repeated integration \cite{cauchy1823, riemann1876}
\begin{align}
\label{eq:fractional_integral}
I^\alpha f(x) = \frac{1}{\Gamma(\alpha)}\int_{a}^{x} (x-t)^{\alpha-1} f(t)\, dt \\
D^\alpha f(x) = I^{-\alpha} f(x)
\end{align}
However, for negative $\alpha$ values, the first equation is undefined. One approach to circumvent this issue is to make use of the ceiling function. For instance, to compute \( D^{1.2} \), we can first take the second derivative and then integrate with an order of \(0.8\) . This allows us to compute fractional derivatives based on negative $\alpha$ values
\begin{equation}
D^\alpha f(x) = \frac{d^{[\alpha]}}{dx^{[\alpha]}} (I^{[\alpha] - \alpha} f(x))
\label{eq:D_operator_rl}
\end{equation}
By substituting $I$, we derive the Riemann–Liouville (RL) fractional derivative \cite{liouville1855}
\begin{equation}
D^\alpha f(x) = \frac{1}{\Gamma([\alpha] - \alpha)} \frac{d^{[\alpha]}}{dx^{[\alpha]}} \int_{a}^{x} (x-t)^{[\alpha] - \alpha-1} f(t)\, dt
\end{equation}
The properties of the Riemann–Liouville (RL) derivative have been extensively studied, and numerous other formulations have been proposed. One major drawback of this specific formulation is the handling of constant functions, i.e. $D^\alpha c = \frac{cx^{-\alpha}}{\Gamma(1-\alpha)} \neq 0$.
To circumvent this, we can reverse the order in which differentiation and integration are applied in \eqref{eq:D_operator_rl} 
\cite{oldham1974fractional}
\begin{equation}
D^\alpha f(x) = I^{[\alpha] - \alpha} (\frac{d^{[\alpha]}f}{dx^{[\alpha]}}(x))
\label{eq:D_operator_caputo}
\end{equation}
\begin{equation}
D^\alpha f(x) = \frac{1}{\Gamma([\alpha] - \alpha)} \int_{a}^{x} (x-t)^{[\alpha] - \alpha-1} \frac{d^{[\alpha]}f}{dx^{[\alpha]}} (t)\, dt
\end{equation}
This is called the Caputo derivative, and while it is widely used in many applications, it may be too computationally expensive when applied to optimization tasks. A more direct approach is to generalize the finite difference form obtaining the Grünwald–Letnikov derivative \cite{oldham1974fractional}
\begin{equation}
D^\alpha f(x) = \lim_{h \to 0} \frac{1}{h^\alpha} \sum_{k=0}^{\left\lfloor \frac{x}{h} \right\rfloor} (-1)^k \binom{\alpha}{k} f(x - kh)
\end{equation}

\subsection{Geometric interpretation}
One way to conceptualize the derivative is as an approximation of a linear map near a given point of a function. Take for example $f: \mathcal{R}^2  \rightarrow \mathcal{R}^2$
\begin{equation}
f(x, y) =  (x^2 - y^2, 3xy)
\end{equation}
% \begin{figure}[H]
% \centering
% \subfigure[]{\includegraphics[width=0.3\textwidth, height=3cm]{images/fig1/f1.eps}}
% \subfigure[]{\includegraphics[width=0.3\textwidth, height=3cm]{images/fig1/f2.eps}}
% \subfigure[]{\includegraphics[width=0.3\textwidth]{images/fig1/f0.eps}}
% \caption{Representations of $f$:
% \textbf{(a)} first dimension 3D plot
% \textbf{(b)} second dimension 3D plot
% \textbf{(c)} grid transform in 2D
% }%
% \end{figure}
Given the Jacobian matrix defined as:
\begin{equation}    
\mathbf{J}_f = \begin{pmatrix}
2x & -2y \\
3y & 3x
\end{pmatrix}
\end{equation}
we define a transformation \( T_{\mathbf{J}_f}(x) = \mathbf{J}_f x \). In this context, \( T \) serves as the best linear approximation to the function \( f \) at a given point \( x \).
However when defining the fractional Jacobian matrix:
\begin{equation}    
\mathbf{J}^\alpha_f = \begin{pmatrix}
x^{-\alpha}(\frac{2x^2}{\Gamma(3-\alpha)}-\frac{y^2}{\Gamma(1-\alpha)}) & y^{-\alpha}\left(\frac{x^2}{\Gamma(1-\alpha)}-\frac{2y^2}{\Gamma(3-\alpha)}\right) \\
\frac{3yx^{1-\alpha}}{\Gamma(2-\alpha)} & \frac{3xy^{1-\alpha}}{\Gamma(2-\alpha)}
\end{pmatrix}
\end{equation}
this changes. The transformation \( T_{\mathbf{J}^{\alpha}_f} \) provides a linear approximation only when \( \alpha = 1 \). As the difference between \( \alpha \) and 1 increases, the non-linear behavior in \( \mathbf{J}^{\alpha}_f \) becomes more pronounced, deviating further from the linear approximation. While losing linear properties, the fractional Jacobian evolves to follow the global curvature of the function more closely. This enables a more faithful approximation of the objective function, potentially capturing more complex behaviors

\begin{figure}[H]
\centering
\subfigure[]{\includegraphics[width=0.3\textwidth]{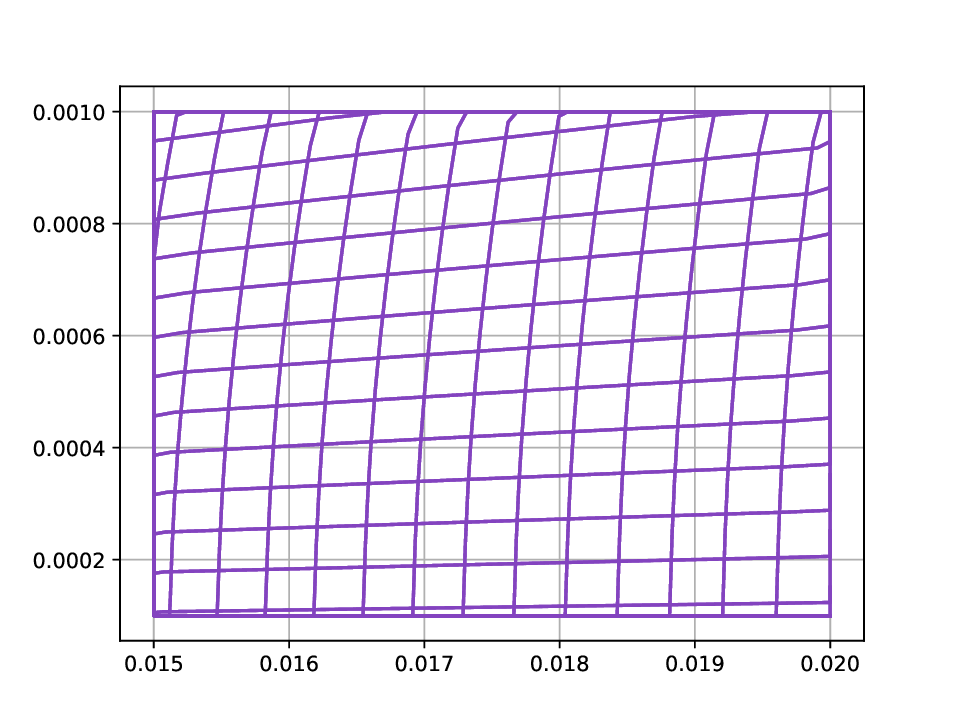}}
\subfigure[]{\includegraphics[width=0.3\textwidth]{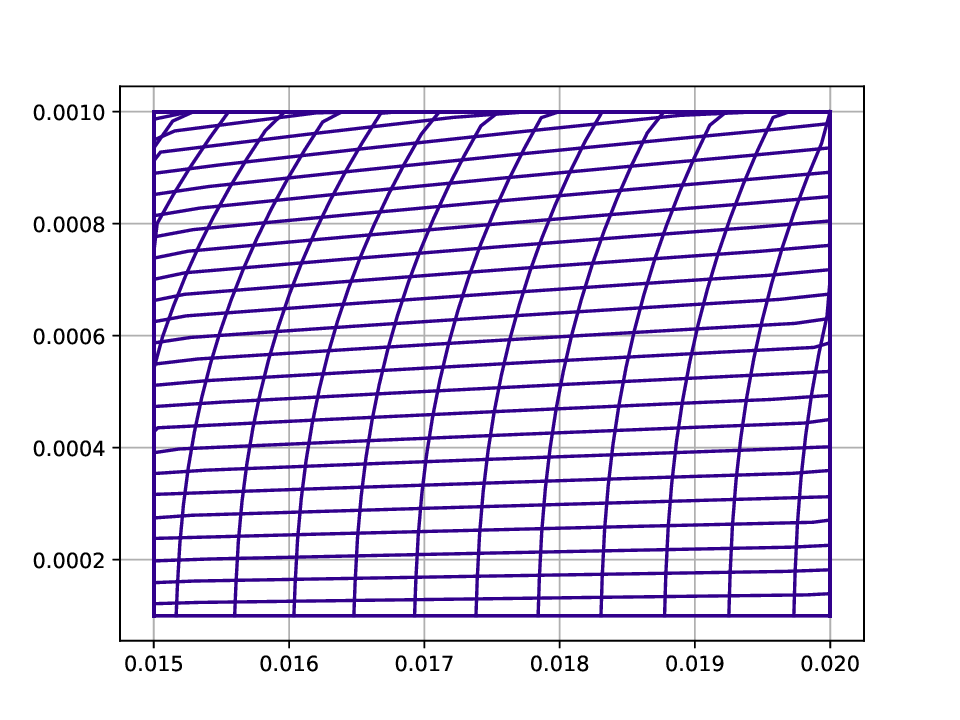}}
\subfigure[]{\includegraphics[width=0.3\textwidth]{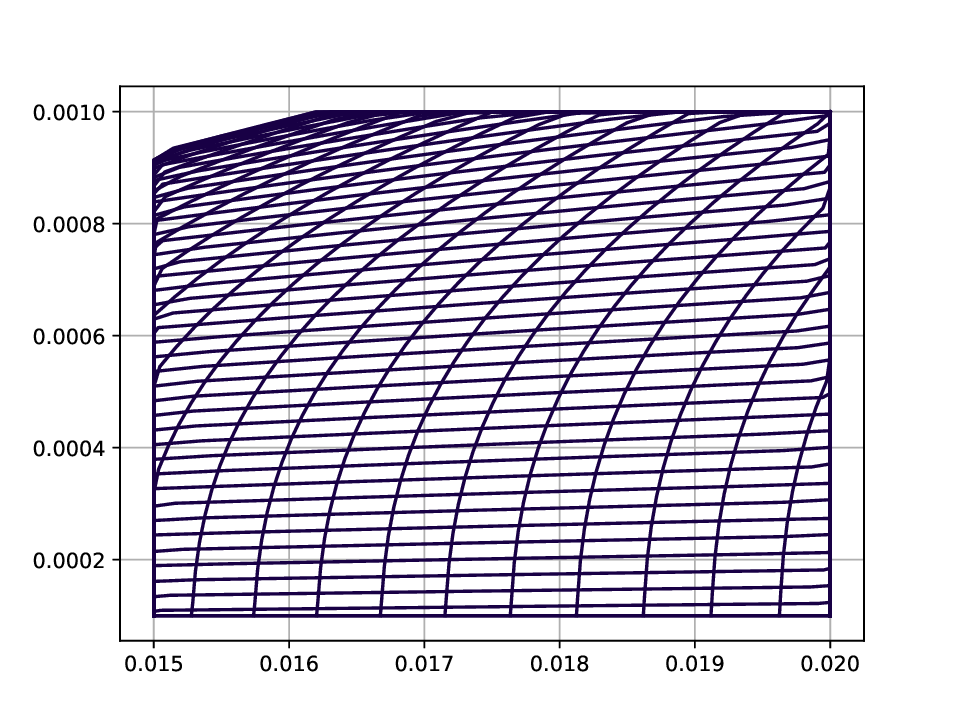}}
\caption{Grid transform near origin for
\textbf{(a)} $T_{\mathbf{J}_f}$
\textbf{(b)} $T_{\mathbf{J}^{1.2}_f}$
\textbf{(c)} $T_{\mathbf{J}^{1.25}_f}$ \\
}%
\end{figure}
\section{Methodology}
\subsection{Meta-learning on classical functions} \label{methodology:clasical_functions}
We start by learning to optimize on a collection of classical functions\footnote{https://www.sfu.ca/~ssurjano/optimization.html}. We consider each function to be parameterized by a state vector $\mathbf{X}_t$. We train a neural network $\mathcal{F}_{\theta}$, that takes as input the current state, normalized gradients, magnitudes of the gradient and Fourier features $\gamma(\mathbf{X}_t)$ \cite{tancik2020fourier}, and outputs the order of the fractional derivative, $\alpha_t \in [0, 1]$ and the magnitude of the update step, $\eta_t$, which are then used to compute the next state $\mathbf{X}_{t+1}$
\begin{equation}
F_{\theta}(\mathbf{X}_{t}, \frac{\nabla f(\mathbf{X}_{t})}{|\nabla f(\mathbf{X}_{t})|}, |\nabla f(\mathbf{X}_{t})|, \gamma(\mathbf{X}_t)) = (\alpha_t, \eta_t)
\end{equation}
Afterwards we can compute the next target function state based on the predicted order and magnitude
\begin{equation}
\label{eq:update_step}
\mathbf{X}_{t+1} =  \mathbf{X}_t - \eta D^\alpha f(\mathbf{X}_t)  
\end{equation}
where $D^\alpha$ is approximated by first order truncated Taylor expansion, 
\begin{equation}
\label{eq:tylor}
D^\alpha f(\mathbf{X} + \Delta \mathbf{X}) =  \binom{\alpha}{0} \frac{1}{\Gamma(1- \alpha)}f(\mathbf{X}) + \binom{\alpha}{1} \frac{1}{\Gamma(\alpha)} \Delta\mathbf{X} \frac{\partial f}{\partial \mathbf{X}}.
\end{equation}
Then using AdamW \cite{adamw} we train the neural network and optimize the objective function
\begin{equation}
    \mathcal{L}_\theta = \log(f(\mathbf{X}_{t+1})) - \log(f(\mathbf{X}_{t}))
\end{equation}
We train this neural network in 2 regimes: 
\begin{itemize}
\item \textbf{with supervision} at each step sample from \textbf{all} the classical functions and merge them in one single batch.
\item \textbf{without supervision} at each step sample from classical functions \textbf{except} the target function.
\end{itemize}
During training, we experimented with using different functions in every batch to help our network generalize better. This indeed helped, albeit only after including Fourier features \cite{tancik2020fourier}. Without them it was difficult for the network to learn high frequency details in low level dimensions.

\subsection{Chaotic systems} \label{methodology:chaotic_systems}

As there are more than one definitions of chaotic systems, it is easier for us to describe chaotic systems by their properties. Most importantly for our work is the sensitivity of the system to the initial conditions. Take for example the target function\footnote{https://lukemetz.com/exploring-hyperparameter-meta-loss-landscapes-with-jax/}
\begin{equation}
\label{eq:target_function}
    f(x) = \log(x^2 + 1 + \sin(3x)) + 1.5
\end{equation}
Applying classical gradient descent on $f$ exhibits chaotic behavior (shown in fig.~\ref{fig:fgf_results}) as a small change in the initial condition (e.g. learning rate, momentum) induces a big change in the final value $x$ over a large enough horizon. 
We can try to apply our proposed fractional gradient descent, but computing the fractional gradient already presents significant challenges from a numerical perspective. We presented a way to approximate it using a first order Taylor expansion in \eqref{eq:tylor}, but, for more complex problems, this is not accurate enough. Therefore, in the context of chaotic systems we propose to analyze the Fractional Gradient Flow (\textbf{FGF}) equation by extending \eqref{eq:update_step} in continuous time
\begin{equation}
    \frac{d^\alpha\mathbf{X}(t)}{dt} = - \nabla f(\mathbf{X}(t)) 
\end{equation}
There are several methods available to approximate fractional differential equations (FDEs) \cite{liang2014, deniz2020, liangWY2020}. In \cite{deniz2020}, the following approximation is proposed:
\begin{equation}
    \mathbf{X}_{k+1} = -\eta ^ \alpha \nabla f(\mathbf{X}_k) + \left( \alpha \mathbf{X}_k - \frac{\alpha(\alpha-1)}{2} \mathbf{X}_{k-1} + \hdots + \frac{\alpha(\alpha-1) \hdots (\alpha - k)}{(k+1)!} (-1) ^ k \mathbf{X}_0 \right)   
\label{eq:fgf}
\end{equation}
This discretization form has a convergence rate of $O(\frac{1}{t^\alpha})$ for locally Lipschitz continuous functions. 
We aim to use this discretization of FGF equations to reproduce the experiment defined in \cite{li2023variance}, where the parameters of a Lorenz system need to be optimized. We compare against various other gradient estimators methods and classic backpropagation through time, using the scheme presented in eq.~\ref{eq:fgf}.
The Lorenz system is defined by the following set of three coupled nonlinear  differential equations:
\begin{align}
\frac{dx}{dt} = \sigma (y - x);
\frac{dy}{dt} = x (\rho - z) - y;
\frac{dz}{dt} = x y - \beta z 
\end{align}
where $x$, $y$, and $z$ are functions of time $t$, and $\sigma$, $\rho$, and $\beta$ are parameters that control the behavior of the system. Our goal is to 
optimize for the control parameters $\mathbf{X} = [\log(\sigma), \log(\rho)]$, starting from an initial state $s_0 = (x_0, y_0, z_0) = (1.2, 1.3, 1.6)$

\section{Results}
\subsection{Meta-learning on classical functions}
\label{results:clasical_functions}
We compare our method against a multitude of optimizers: a general purpose learned optimizer (VeLO), classic gradient descent and adaptive methods. For testing, we randomly sample $1000$ starting points from the target function domain and report the convergence rates for a maximum of $100$ steps of optimization. We consider a solution to be converged if the value of the function at that point is at a maximum distance of $\epsilon=10^{-3}$ from the global minimum. We also report the average number of steps to reach the solution. For learning rate based methods we search over learning rates between $10^{-1}$ and $10^{-6}$ and present the run with the best results.
\begin{table*}[!htbp]
    \small
    \centering
    \begin{tabular}[t]{l||cc}
    % \hline
    \textbf{Optimizer}  & {Convergence Rate} & {Truncated trajectory length} \\
    \hline
    (1) GD & $0.60\%$ & $994.03$ \\
    \hline
    (2) Adam & $1.60\%$ & $985.15$ \\
    \hline
    (3) AdamW & $1.60\%$ & $985.16$ \\
    \hline
    (4) RMSProp & $3.50\%$ & $966.09$ \\
    \hline
    (5) Adafactor & $1.10\%$ & $989.27$ \\
    \hline
    (6) Adagrad & $0.90\%$ & $991.48$ \\
    \hline
    (7) VeLO \cite{velo} & $0.30\%$ & $997.05$ \\
    \hline
    (8) Ours \\w supervision & $99.20\%$ & $12.156$ \\
    \hline
    (9) Ours \\w/o supervision & $71.80\%$ & $321.37$ \\
    \hline
    \end{tabular}
    \caption{\small Performance analysis on Rosenbrock 2D: our trained neural network optimizer predicts updates that closely resemble those of a second-order method. This behavior aligns with expectations for learned optimizers. (8) Our optimizer trained only on the target function with supervision (9) Our optimizer trained on other classical functions \textbf{except} the target function.}
\label{tbl:rosenbrock_function}
\end{table*}
\subsection{Chaotic systems} \label{results:chaotic_systems}
\begin{figure}[t]
\centering
\subfigure[]{\includegraphics[width=0.45\textwidth]{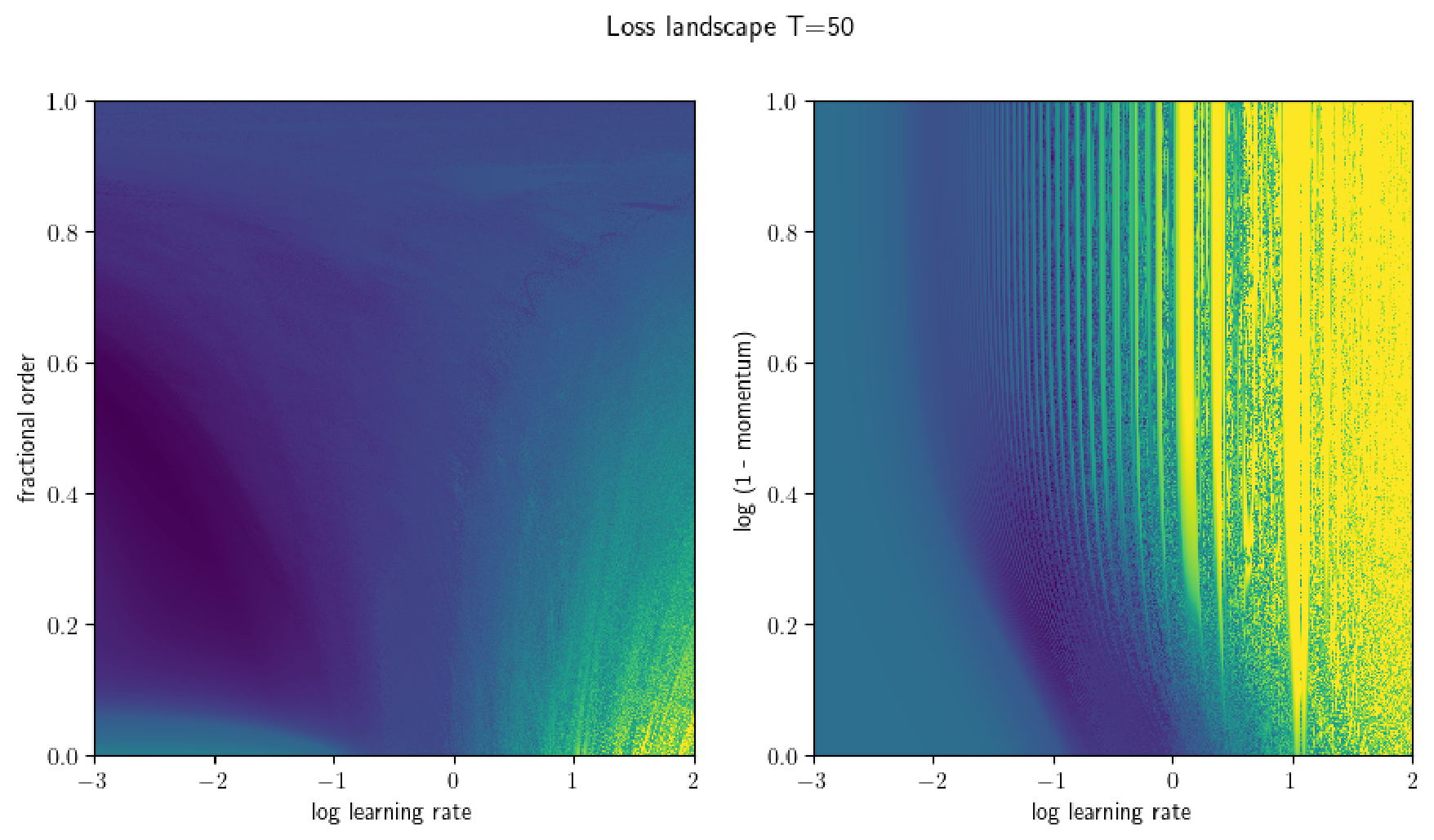}}
\subfigure[]{\includegraphics[width=0.45\textwidth]{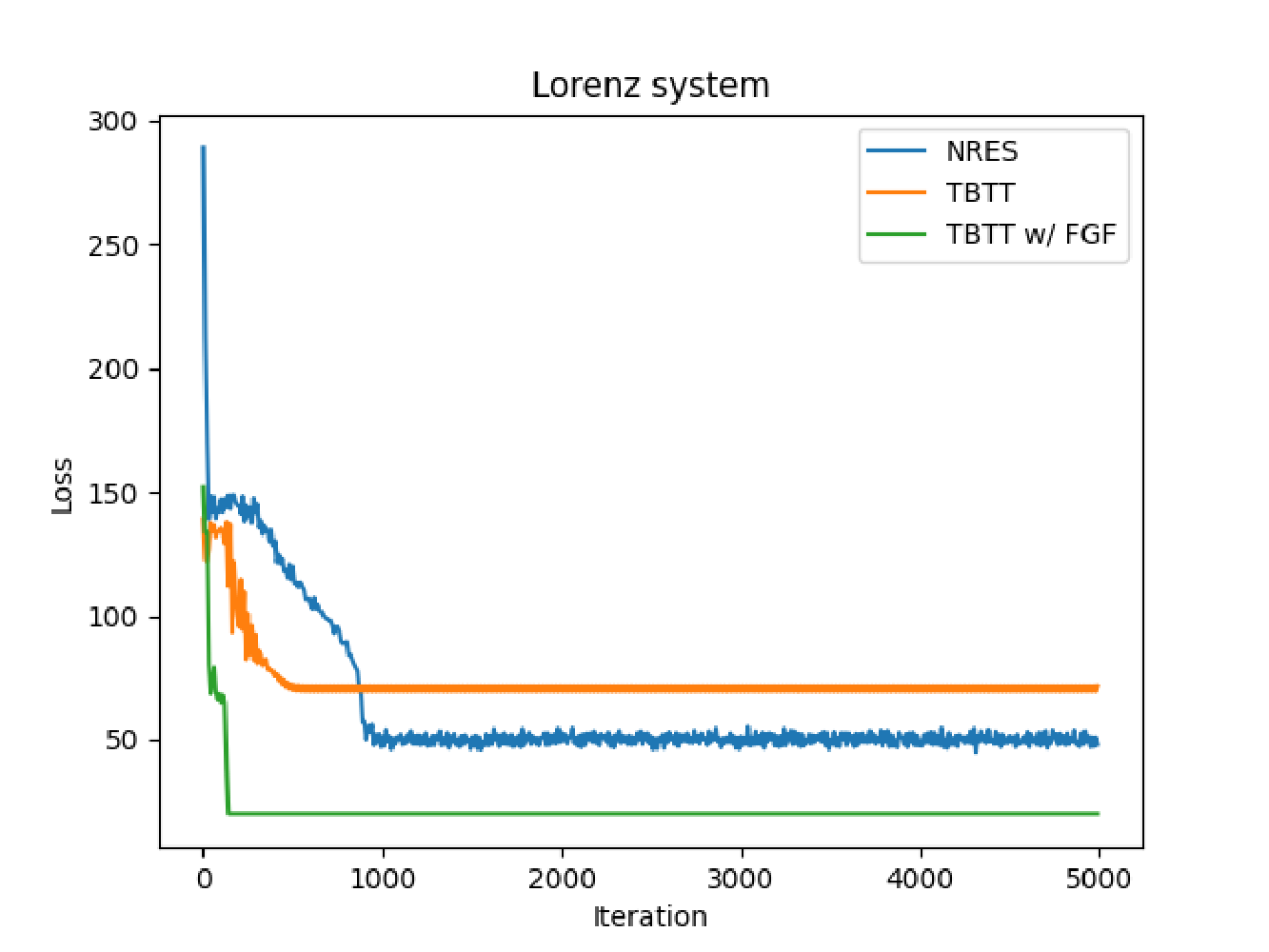}}
\caption{a) Loss surface for the function from eq.~\ref{eq:target_function} (left) classical gradient descent searching over momentum decay and learning rates; (right): our FGF method searching over fractional orders and learning rates. 
b) Loss convergence for the Lorenz optimization problem defined in \cite{li2023variance} comparing NRES and TBTT, with and without the FGF discretization scheme.
}%
\label{fig:fgf_results}
\end{figure}

We conducted two experiments, as described in section \ref{methodology:chaotic_systems}. The first one involves a comparison of the loss landscape for the function in \eqref{eq:target_function}, between classical gradient descent and the FGF method. This experiment can be seen in fig.~\ref{fig:fgf_results}(a).

The second experiment regarding the optimization of the Lorenz system, compares the update step from \eqref{eq:fgf} with the update step generated by classical backpropagation through time (\textbf{TBTT}) and the update step generated by the gradient estimator presented in \cite{li2023variance}, \textbf{NRES}. Changing the update step from classical gradient descent to fractional gradient descent makes \textbf{TBTT} perform the best, as it can be seen in fig.~\ref{fig:fgf_results}(b). This approach converges faster and is more stable. Although this is a toy problem, we believe that there may be potential for such techniques in real-world applications. Currently, due to the chaotic nature of TBTT, Evolutionary Search gradients are used, such as NRES. These methods, although stabilize the exploration of the loss landscape, suffer from poor convergence rates.

\section{Limitations}
We investigated various forms of the fractional derivative, which show promising results in small-dimensional problems such as classical optimization tasks and traditional chaotic systems. Although their application is currently limited in meta-training, we believe that further research could make fractional calculus a viable approach. We believe that one of the problems of extending this work to higher-dimensional problems, is that a single fractional order might not accurately describe all the dimensions. In recent works, Transformers were used to overcome the problem of dimensionality in learned optimization \cite{gartner2023transformer}, but the inference time is slow. We aim to take advantage of other advances in the field, such as faster attention mechanisms \cite{dao2023flashattention2} or SSM \cite{gu2022efficiently} and make this approach more feasible in the future. 
In meta-learning, the inner and outer dynamics of the system differ. In this work we focused solely on the inner dynamics which sometimes behaves akin to a chaotic system. Further investigation is needed to extend this work also to the outer dynamics.\\
\textbf{Acknowledgments:} This work was supported in part by project CNCS-UEFISCDI (PN-III-P4-PCE-2021-1959). The authors thank Andrei Zanfir and Mykhaylo Andriluka for their insightful feedback throughout various stages of this project.
\bibliography{main}
%\newpage
%\clearpage
%\appendix

%\section{More}
%\input{appendix}

\end{document}